\documentclass[letterpaper,conference]{IEEEtran} 
\usepackage{mathptmx} %

\newcommand{\ignore}[1]{}
\usepackage{fancyhdr}
\usepackage[normalem]{ulem}
\usepackage[hyphens]{url}

\usepackage{amsmath}
\usepackage{graphicx}
\usepackage{xcolor}
\usepackage{subcaption}
\usepackage{xspace}
\usepackage{bbold}
\usepackage{pbox}
\usepackage[rightcaption]{sidecap}
\usepackage{flushend}
\sidecaptionvpos{figure}{c}

\definecolor{Awesome}{rgb}{1.0, 0.08, 0.58}

 \fancypagestyle{firstpage}{
   \fancyhf{}
 \setlength{\headheight}{50pt}

}

\usepackage{xstring}

\makeatletter
\def\@separatefourdigits#1#2{%
  \StrLen{#1}[\length]%
  \ifnum\length<5\relax#1\else%
     \bgroup
     \StrRight{#1}{4}[\lowerfour]%
     \StrGobbleRight{#1}{4}[\remaining]%
     \@separatefourdigits{\remaining}{#2}#2\lowerfour%
     \egroup
  \fi}

\def\formatnumber#1#2{\StrDel{#1}{ }[\aux]%
\@separatefourdigits{\aux}{#2}%
}

\def\formatnumberwithspaces#1{\formatnumber{#1}{ }}
\makeatother

\begin{document}

\title{Dynamic Stripes: Exploiting the Dynamic Precision Requirements of Activation Values in Neural Networks}

\author{
Alberto Delmas,
Patrick Judd,
Sayeh Sharify
\& Andreas Moshovos\\
Department of Electrical and Computer Engineering, University of Toronto\\ 
\texttt{\{delmasl1, juddpatr, sayeh, moshovos\}@ece.utoronto.ca} \\ 
}

\maketitle
\thispagestyle{firstpage}

\pagestyle{plain}

\begin{abstract}
Stripes is a Deep Neural Network (DNN) accelerator that uses bit-serial computation to offer performance that is proportional to the fixed-point precision of the activation values. 
The fixed-point precisions are determined a priori using profiling and are selected at a per layer granularity. 
This paper presents Dynamic Stripes, an extension to Stripes that detects precision variance at runtime and at a finer granularity. 
This extra level of precision reduction increases performance by 41\% over Stripes.

\end{abstract}

\section{Introduction}

Our previously described Stripes~\cite{stripes-CAL} accelerator exploited the variable precision requirements of deep learning neural networks to improve performance and energy efficiency. 
In the previously disclosed design, the hardware expected that prior to processing each layer, the precision required by that layer would be communicated by the software. 
These per layer precisions were thus not adjusted at runtime to reflect any additional reduction in precision that may be possible for each layer or even at a smaller granularity.
However, the underlying compute units are capable of exploiting precisions on a much finer granularity than a layer. 
In the described implementation, each chip comprised 16 tiles, each processing 16 filters and 16 weights (synapses) per filter. 
A set of 256 activations where broadcast to all tiles one bit per cycle. For each layer, the precision of the activations, that is the positions of the most significant and of the least significant bits (MSB and LSB respectively), n\textsuperscript{H} and n\textsuperscript{L} were adjusted per layer. 
However, the precision could easily be adapted at a smaller granularity. 
For example the precision could be adjusted per group of 256 activations that are processed concurrently, or per group of 16 activations that are broadcast to the same column of SIPs in the described implementation.
Similar techniques of dynamic precision reduction for bit-serial computation have been used for signal processing~\cite{xan-low_power_dct-jssc2000}. 
To the best of our knowledge this is the first application of this technique to neural networks.

\section{Simplified Example}
\begin{figure}[htb!] %
\centering
\includegraphics[scale=0.26]{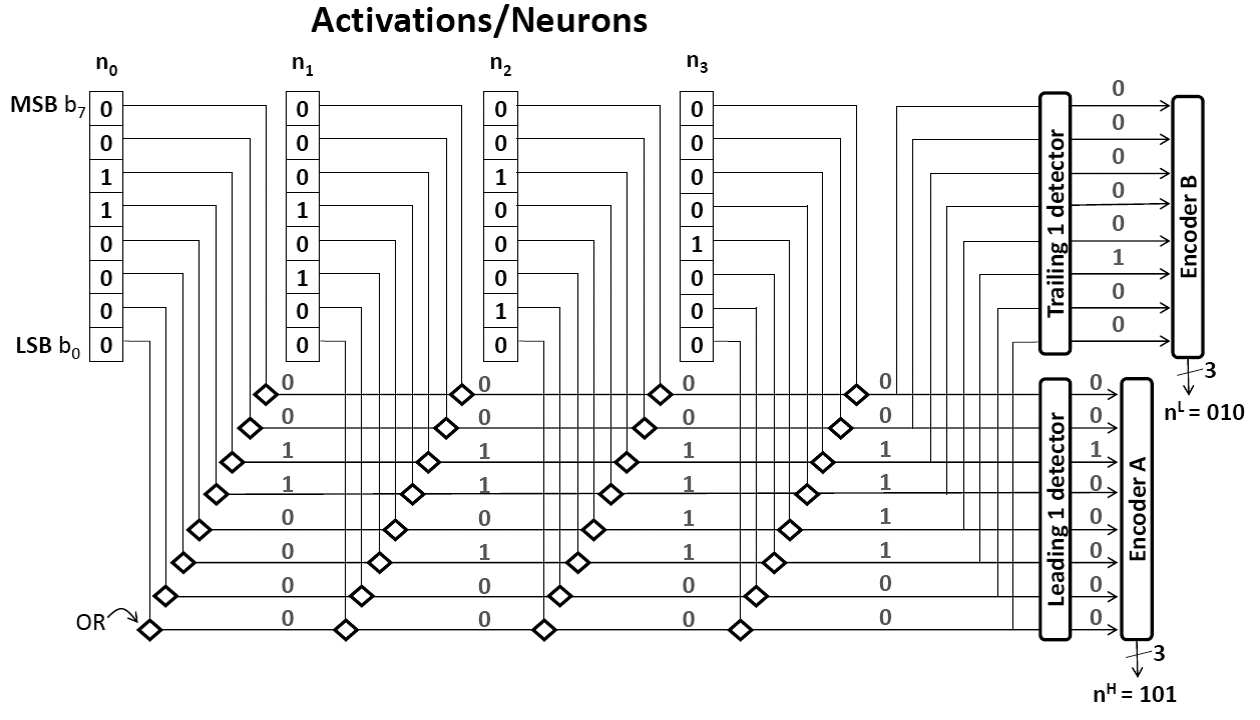}
\caption{An Example Illustrating the Runtime Detection of the Precision Necessary to Process a Group of Activations.}
\label{fig:fig1}
\end{figure}

Figure~\ref{fig:fig1} shows how the detection of n\textsuperscript{H} and n\textsuperscript{L} can be done for an example where a group of four activations, each originally expressed in 8 bits of precision. Without the mechanism to detect the precision at runtime, it could have been that he precision used to process this group of activations would be 7 bits since the precisions have to be determined during a prior profiling run and for the whole layer. This would be the case, if for example, there were some other activation possibly at some other position, that during the profiling runs was determined that it needed to be able take the value \formatnumberwithspaces{0100 0001}. Or, as another example, this would be the case, if there were two activations that during the profiling run it was determined that they needed to be able to take the values of \formatnumberwithspaces{0100 0000} and \formatnumberwithspaces{0000 0001} respectively for accuracy to be maintained as desired.
However, during any run, the activations will take values that depend on the given input to the network (or prior inputs in the case of, for example, recurrent neural networks). As the figure shows, the specific values for this group of four activations can all be represented using just 4 bits. This is because among all four values, the highest bit position a 1 appears is at position 5 (counting from 0), e.g, in n\textsubscript{0}, whereas the lowest bit position where a 1 appears is at position 1 in n\textsubscript{1}. %
 The figure shows the network calculating a set of signals ORj, one per input bit position. The ORj signals can be calculated using a set of cascaded OR gates which in the figure are shown as diamonds. The n\textsubscript{H} detection block accepts these ORj signals as input and performs a leading “bit that is 1”, or simply “leading 1” detection, to identify the position of n\textsubscript{H}.  The result of this block is a 1-hot encoding of n\textsubscript{H}. Since this information has to be communicated to the processing tiles, an encoder block can be used as shown to convert this encoding into a number in binary. This is the function of the offset encoder. The n\textsubscript{L} detection uses an identical block as the n\textsubscript{H} block with the priority of the ORj inputs reversed. This in effect calculates a trailing “bit that is 1” detector. The figure shows an example of detecting n\textsubscript{H} and n\textsubscript{L} for a set of neuron values. Since the input neurons use 8 bits, the two offsets are encoded using 3 bits each. 
To process this group of neurons, the dispatcher will send n\textsubscript{H} as the starting offset. The units will decrement this offset every subsequent cycle. The dispatcher will signal the last cycle of processing for this group when the current offset becomes equal to n\textsubscript{L}. This can be done with an extra wire which signals the end of processing each neuron group. Assuming that processing starts at the n\textsubscript{H} bit, position, a counter keeps track of the current bit position being broadcast and a comparator sets the end of group signal when we arrive at n\textsubscript{L}.  

While in our example there are only four neurons that are being processed concurrently and as a single group, in practice there may be more neurons and more than one group. In such a case, once a group of neurons is processed the corresponding neuron lanes can be made to wait for all other neuron lanes to finish before advancing to the next group of neuron values. Alternatively, the dispatcher and the synapse buffer can be modified to support per neuron group accesses at the expense of additional area and memory bandwidth.

\section{Design} 
While in the previous section we considered a case where only four activations forming a single group are processed concurrently, other implementations may choose to process more activations an in multiple groups. In the example implementation described in Stripes, 256 activations are processed in concurrently. They are broadcast to 16 tiles, where each tile processes 16 filters, and 16 weights per filter for a total of 256 weights per tile. Each tile comprises a grid of 16x16 Serial Inner-Product units, or SIPs. The SIPs along the same row process the same set of 16 weights, while the SIPs along the same column process the same set of 16 activations. Accordingly, in this implementation the precision p, specified as (n\textsuperscript{H},n\textsuperscript{L}) can dynamically adjusted for the whole group of 256 activations that are processed concurrently or it can be determined separately for each group of 16 activations that are processed concurrently by each column of SIPs. 

\begin{figure}[htb!] %
\centering
\includegraphics[scale=0.30]{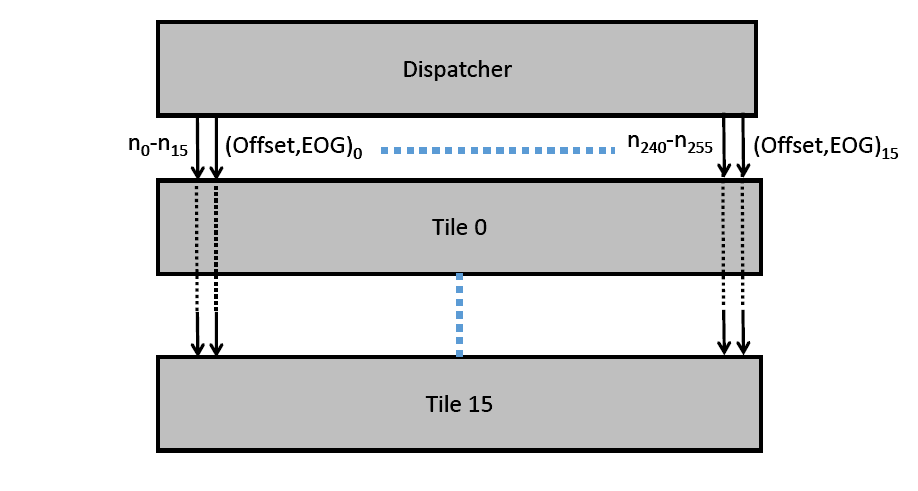}
\caption{A Dynamic Stripes Overview of Dispatcher and Processing Titles.}
\label{fig:fig2}
\end{figure}

Determining the precision necessary for a group of activations can be performed by the dispatcher prior to communicating the activations to the units for processing. For each of the 256 neurons to be sent to the units, $n_i$ where $i \in [0,255]$, and for each bit $n_{i,j}$ $j \in [0,15]$ and assuming that all neurons are positive, the transposer first calculates the logical OR of all bits at the same position: 
$OR_j = \sum_{i=0}^{i=255}n_{ibj} $, and then applies a leading bit that is 1 detector over $OR_j$, $j=[0,16]$ to determine n\textsuperscript{H} the highest bit position where a bit that is 1 appears. Similarly, the transposer uses a trailing bit that is 1 detector to determine n\textsuperscript{L} the lowest bit position where a bit that is 1 appears. To process these neurons, the transposer sends along with the bits also their offset via a set of extra 4 wires. An additional wire indicates the end of processing a neuron group. Assuming that processing starts at the n\textsubscript{H} bit position, a counter keeps track of the current bit position being broadcast and a comparator sets the end of group signal when the units arrive at n\textsuperscript{L}. 
Figure~\ref{fig:fig2} shows an example Dynamic Stripes organization and specifically how the dispatcher communicates the activation bits and precision info to 16 processing tiles. In this configuration the 256 activations are grouped in 16 subgroups, each of 16 activations. For each subgroup the dispatcher sends the current offset along with an End of Group (EOG) signal. Assuming a baseline precision of 16 bits, the offset will require 4 wires and the EOG one for a total of 5 extra wires per subgroup of 16 activations. Other groupings are possible.

\subsection{Modified Serial Inner-Product Unit}
\begin{figure}[htb!] %
\centering
\includegraphics[scale=0.50]{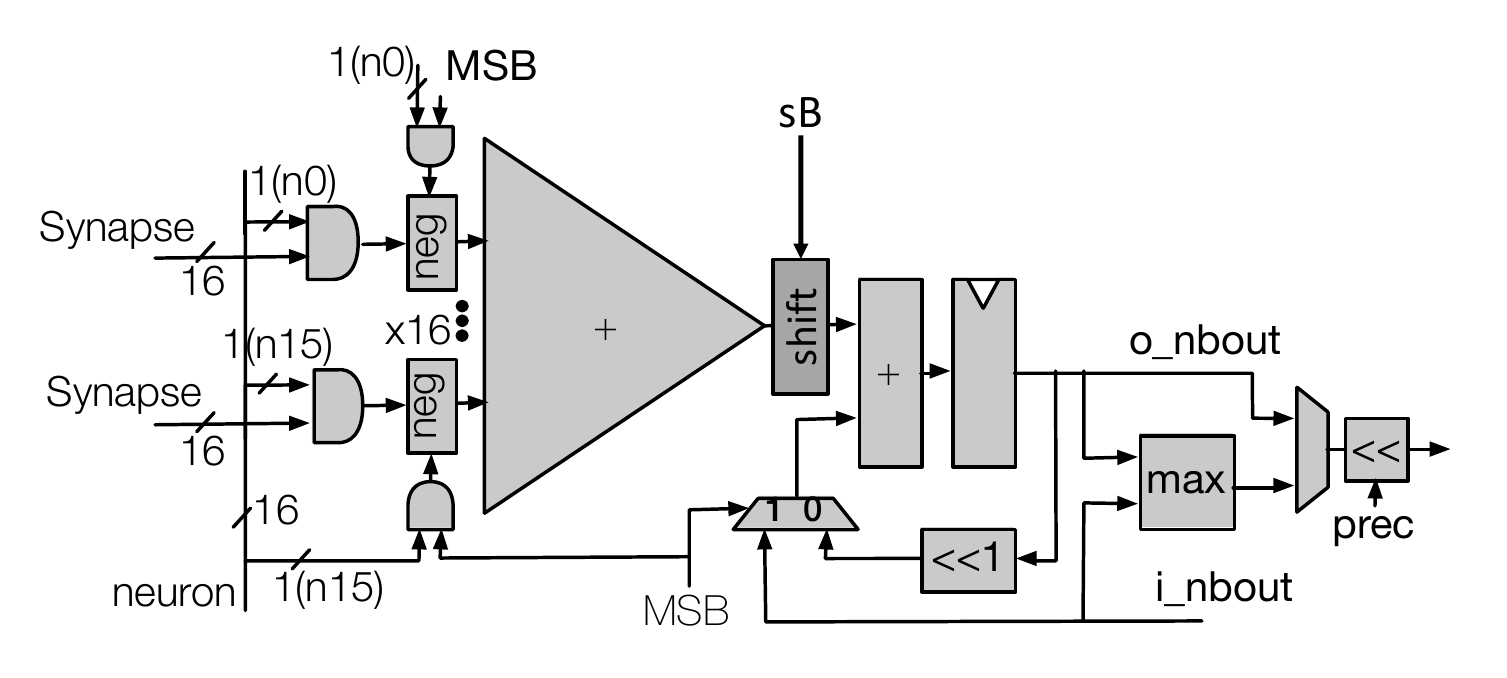}
\caption{Modified SIP.}
\label{fig:fig3}
\end{figure}

Figure~\ref{fig:fig3} shows an implementation of the modified Serial-Inner Product Unit. An additional shifter appears at the output of the adder tree. This shifter, through the control signal sB, allows the output of the adder tree to be adjusted to the right bit position so that it can be accumulated with the running sum which is kept at the accumulator A.

\section{Evaluation: Performance}

We have evaluated the performance improvement possible with Dynamic Stripes for a set of modern Image Classification Convolutional Neural Networks. Table~\ref{tab:t1} shows the resulting performance improvements of an equivalently configured DaDianNao~\cite{DaDiannao} accelerator. Performance measurements were obtained via our cycle-accurate simulator which was extended to support dynamic detection of activation precisions. For this set of results, we first set the precisions per layer as in the original Stripes design and then precision was dynamically detected at runtime per subgroup of 16 activations being processed concurrently. All 16 subgroups belonging to each group of 256 activations that were being processed concurrently had to finish processing prior to advancing to the next set of 256 activations. The Dynamic Stripes design improves performance over the baseline Stripes design. It lags compared to the Pragmatic~\cite{DBLP:journals/corr/AlbericioJDSM16} design that skips zero-bits instead of trying to exploit precision alone.

\begin{table}[tbp]
\centering
\begin{tabular}{|l|l|l|}
\hline
\textbf{Network}   & \textbf{vs STR}   & \textbf{vs DaDN}    \\ \hline
AlexNet   & 1.29x    & 2.81x      \\ \hline
NiN       & 1.30x    & 2.39x      \\ \hline
Googlenet & 1.56x    & 2.64x      \\ \hline
VGG\_2    & 1.43x    & 3.16x      \\ \hline
VGG\_S    & 1.62x    & 3.28x      \\ \hline
VGG\_19   & 1.27x    & 1.71x      \\ \hline
GeoMean   & 1.41x    & 2.61x      \\ \hline
\end{tabular}
\captionof{table}{Speedup} \label{tab:t1} 
\end{table}

\section{A Novel way to adjust precision with Pragmatic}

In the Pragmatic engine which is was motivated by Stripes, performance depends on the number of activation bits that are 1. This engine opens up the opportunity for value based optimizations for performance and energy efficiency where the activation values are adjusted on the fly to reduce the number of bits that are 1. For example, we can determine  a per layer threshold for the number of most significant powers of 2 (MSP2) that ought to be kept to maintain accuracy. For example, if an activation has a value \formatnumberwithspaces{1010 0101} and MSP2 is 3, the activation can be converted to \formatnumberwithspaces{1010 0100}, whereas if MSP2 was 2 it would be converted to \formatnumberwithspaces{1010 0000}. With either MSP2 threshold, an activation of \formatnumberwithspaces{0000 0101} would be represented exactly. This representation is different than a fixed-point and a floating point representation. We have experimented with such an optimization and found that it greatly improves performance for AlexNet. Specifically, the method followed is: 1. Get precision profiles via gradient descent to determine the number of 1-bits required for 100\% accuracy. For Alexnet this gives 5,5,5,3,3, as opposed to 9,8,5,5,7 when using a fixed number of bits. 2. Take the n most significant 1-bits of every neuron, where n is the precision profile obtained in step 1, and then run that through Pragmatic.  Performance over DaDianNao becomes 3.91x with the 2b-1R Pragmatic configuration (the tiles remain the same, only the dispatcher needs to be modified to stop transmitting oneffsets after a threshold). By comparison, sending all oneffsets resulted in a performance improvement of 3.26x. With full range shifters, the performance with this technique becomes 4.83x which exceeds that of the improved encoding technique.

\Urlmuskip=0mu plus 1mu\relax
\bibliographystyle{ieeetr}
\bibliography{ref}

\end{document}